\documentclass{article}

\usepackage[preprint]{neurips_2019}
\usepackage[utf8]{inputenc} % allow utf-8 input
\usepackage[T1]{fontenc}    % use 8-bit T1 fonts
\usepackage{hyperref}       % hyperlinks
\usepackage{url}            % simple URL typesetting
\usepackage{booktabs}       % professional-quality tables
\usepackage{amsfonts}       % blackboard math symbols
\usepackage{nicefrac}       % compact symbols for 1/2, etc.
\usepackage{microtype}      % microtypography
\usepackage{amsmath}
\usepackage{graphicx}
\usepackage{subcaption}
\usepackage{authblk}

\title{Adaptive Symmetric Reward Noising for Reinforcement Learning}

\graphicspath{ {./images/} }
\author[1]{Refael Vivanti}
\author[1]{Talya D. Sohlberg-Baris}
\author[1]{Shlomo Cohen}
\author[1]{Orna Cohen}
\affil[1]{Rafael ltd}

\begin{document}

\maketitle

\begin{abstract}
Recent reinforcement learning algorithms, though achieving impressive results in various fields, suffer from brittle training effects such as regression in results and high sensitivity to initialization and parameters. We claim that some of the brittleness stems from variance differences, i.e. when different environment areas – states and/or actions – have different rewards variance.
This causes two problems: First, the "Boring Areas Trap" in algorithms such as Q-learning, where moving between areas depends on the current area variance, and getting out of a boring area is hard due to its low variance. Second, the "Manipulative Consultant" problem, when value-estimation functions used in DQN and Actor-Critic algorithms influence the agent to prefer boring areas, regardless of the mean rewards return, as they maximize estimation precision rather than rewards. This sheds a new light on how exploration contribute to training, as it helps with both challenges. 	
Cognitive experiments in humans showed that noised reward signals may paradoxically improve performance. We explain this using the two mentioned problems, claiming that both humans and algorithms may share similar challenges. Inspired by this result, we propose the Adaptive Symmetric Reward Noising (ASRN), by which we mean adding Gaussian noise to rewards according to their states' estimated variance, thus avoiding the two problems while not affecting the environment's mean rewards behavior. We conduct our experiments in a Multi-Armed-Bandit problem with variance differences. We demonstrate that a Q-learning algorithm shows the brittleness effect in this problem, and that the ASRN scheme can dramatically improve the results. We show that ASRN helps a DQN algorithm training process reach better results in an end to end autonomous driving task using the AirSim driving simulator. 
\end{abstract}

\section{Introduction and related work}

Reinforcement Learning (RL) has shown great success during the last few years in achieving superhuman performance in many areas. However, most state-of-the-art algorithms have major problems, such as a need for high sample complexity, and brittle convergence showing problems  such as regression in results and high sensitivity to initialization and parameters. The current trend in reinforcement learning research is to treat such problems as if they stem from the training process arriving at a local-minimum [1]. Therefore, the typical solutions are multiple runs with different random initializations or with slightly different parameters, and different exploration strategies.\\
However, these solutions introduce new challenges. Multiple initializations, combined with the high sample complexity in RL, require considerable computing resources. Exploration strategies interfere with the behavior of the regular policy. Extrinsic exploration, such as epsilon-greedy methods [2], enforce different actions than those suggested by the policy. In cases where the epsilon parameter decreases in order for the real policy to take control over time, it is common that the training agent retreats to a less successful policy when there is little exploration due to a low epsilon. \\
Recent works introduce intrinsic exploration, by learning an intrinsic curiosity model (ICM), and the curiosity or states-entropy are either used as the sole reward system [3] or combined with the true rewards [4]. However, changing the reward system during training can produce a policy which behaves significantly different from the desired one. In many tasks, low-entropy states have large rewards, and a well-trained curiosity-driven agent will gain comparably low scores. Specifically, intrinsic exploration methods assume that agent-failure moves the player to the same game start, which will cause the curiosity-driven agent to prefer avoiding low entropy failures. However, other challenges such as control tasks may have random initializations, and in those cases the curiosity-driven agent will consider, as the song says, that suicide is painless, since it brings up many changes. [5]\\
Several works discuss noise in RL. Usually noise is considered as a negative factor in RL. For example Fox et al. [6] have suggested adding a cost term to penalize deterministic policies for noise taming. Conversely, several works integrate noise as a positive factor. The common  Epsilon-greedy exploration scheme adds noise to the action space [2].  Recently, Plappert et al. [1] have suggested adding noise to the agent's parameters, for more consistent exploration. \\
We have not found a use of reward noising in RL algorithms. Interestingly, Gureckis et al. [7] found in cognitive experiments that noise and uncertainty in reward signals may paradoxically improve human performance. They added four levels of normally distributed noise to rewards in the "Farming on Mars" task, and found that high amount of noise reduces performance as the signal-to-noise ratio is too low, but low amount of noise improves performance. They hypothesize that the noise may have helped to encourage more effective exploration in the task, as inconsistent feedback might rule out simple hypotheses, which were reinforced by relatively low rewards variability, and encourage sampling of alternatives.  \\
Similarly, we claim that some of the brittleness in RL algorithms stems from variance differences in the environment, when different environment areas - states and/or actions - have different rewards variance. This may cause two problems. The first one is the "Boring Areas Trap" in algorithms such as Q-learning, where moving between areas depends solely on the variance of the current area, thus getting out of a boring area is hard due to its low variance. The second is the "Manipulative Consultant" problem, when value-estimation functions used in DQN and Actor-Critic algorithms influence the agent to prefer boring areas, regardless of the mean rewards return, as they maximize estimation precision rather than rewards. We explain how exploration can be interpreted as a partial solution to those problems, and that its importance is beyond introducing new options to the agent.\\
Inspired by the cognitive experiments in [7], we propose introducing reward noising to RL. Following their work, the noise should be symmetric so it would not change the mean reward. However we do not want to add noise to already noisy areas, as it can reduce performance. We therefore propose our "Adaptive Symmetric Reward Noising" scheme (ASRN) which adaptively adds noise to the rewards according to their states' estimated entropy. \\
We explore a Multi-Armed-Bandit task with variance differences, which despite the scenario's extreme simplicity, still reproduces the brittleness challenge in Q-learning, the "Manipulative Consultant" and the "Boring Areas Trap" problems, and will examine how our ASRN scheme can help overcoming them. 

\section{Variance differences implications}

\subsection{Variance differences}
Our first observation is that in many environments, different areas -  either states, actions or combinations of the two- have different reward variance. In some cases, the mean reward in low-variance areas might be lower than the mean reward in high-variance areas. We call low-variance-low-mean areas "boring areas", in contrast to high-variance-high-mean areas which we call "interesting areas". Note this is not the usual interpretation of these phrases in other RL literature such as [8]. \\
Examples for boring area might include 'turn a car right in circles' (action), 'stand in place' (action), 'commit suicide when possible' (state-action), or any other areas with low mean reward and low rewards variance. Variance difference can affect value-estimation functions that try to predict the rewards from the state and action, since in low-variance areas the predictions tend to be more accurate.

\subsection{The Boring Areas Trap}
The first problem is when using plain Q-learning algorithm in an environment with variance differences. Suppose there are only two areas, boring and interesting, and the Q-learning algorithm has to decide where to go. When there is a variance in the rewards of the interesting area $r_t^i$, there will also be a variance in the value-estimation of the interesting areas $Q_i^t$, though a considerably lower one. In the rare case of a concentration of low rewards in a short period, $Q_i^t$ will be temporarily lower than the value-estimation of the boring areas $Q_b^t$. At this point the consultant will recommend boring actions, the agent will become boring, and the process enters what we call the "Boring Areas Trap". \\
Once the policy becomes a boring one, it will only sample examples from the boring areas. When this happens, getting back to an interesting policy, requires a \textbf{complementary error}, in which a series of bad boring examples reduces the current boring areas' value-estimation $Q_b^t$ below the (currently low) $Q_i^t$. If $\sigma_b$ is low, the variance of $Q_b^t$ will be very low, the complementary error will hardly occur, and the policy will be boring most of the time. This is why we call it a "trap". In the special case of $\sigma_b=0$ , such as in the case of a suicide-button with a constant reward, or a non-functioning action which does nothing with no reward, there is no exit from the Boring Areas Trap. This type of entering the Boring Areas Trap can happen even to a well-trained agent, if the training process continues. This might be the reason to the retreat in performance phenomenon in RL algorithms.\\
A popular solution commonly used is clipping [9, 10] the influence of a temporal reward on the policy, which can sometimes prevent falling into the Boring Areas Trap by reducing  $Q_i^t$ variance. However, clipping also reduces $Q_b^t$ variance and by doing so interferes with the ability of a policy to get out of this trap. Also, when a very small learning rate $\alpha$ is chosen, $Q_i^t$ is influenced by many rewards, so due to the  central limit theorem, even large $\sigma_i$ will not produce high variance in $Q_i^t$, thus falling into the Boring Areas Trap will rarely happen, unless $\mu_i-\mu_b$ is very small. This limits the choices of $\alpha$ in RL implementations to very small ones, and thus slows convergence due to low learning rate. \newline
Another option to be in the Boring Areas Trap stems from the process initialization parameters. Those may be the initial estimations in the Q table in Q-learning, or the random weights of the neural network in DQN. Since the initial value estimations are distributed i.i.d. to the environment's rewards, for each state which may have an interesting action and a boring action, about half the agents will have $Q_b^0 > Q_i^0$, and they will be trapped in the Boring Areas Trap for this state. For an environment with many such states, the probability for an initialization which is not in the Boring Areas Trap for any state is exponentially low. This may be where the sensitivity to initialization of RL algorithms stems from. 
The Boring Areas Trap bare resemblance to the phenomenon Gureckis et al. found in [7], where the human experiment participants persisted in choosing the lower-rewards option due to it's low variance.

\subsection{The Manipulative Consultant Problem}
The second problem appears when RL algorithms use a value-estimation function for estimating the future rewards. It is used as the main component of DQN methods, or as the critic function in Actor-Critic methods. We call this type of function a "Consultant", as it does not maximize the desired function – the returned rewards – but rather advises the agent regarding the consequences of a specific action given a state. Consultant functions are very popular in RL as they can learn off-policy, i.e. learning from the experience of other policies. This enables learning from the entire accumulated experience, even from episodes performed using an old policy. In recent RL algorithms, the consultant is implemented as a deep neural network which estimates the value of a state and action, because the states space is very large. The network's weights are optimized using a gradient-descent process aiming to minimize a loss function, which measures the estimation accuracy.\\
The main problem we describe here is that the consultant's loss function measures its own accuracy rather than the agent's reward. Therefore, in the presence of variance difference, the consultant's loss will be lower on low-variance areas than on high-variance ones.\\
This creates a conflict of interests: if the consultant produces false results which manipulate the agent towards boring areas, it will predict the outcome more accurately, thus reducing its own loss. As an allegory, we can think of a human investments-consultant who is paid according to the accuracy of his prediction instead of the revenues of the customer, and manipulatively advises the customer to choose low-risk-low-return investments in order to maximize his own profit.
Additionally, since the consultant affects the agent's behavior, if it will suggest only low-variance areas it will be tested only on them. If the suggestion stems from an inaccurate estimation of the high-variance areas, this inaccuracy will not influence the loss as those areas will not be visited.\\
We say a value-estimation function is a "Manipulative Consultant" when it manipulates the estimation values to produce a sub-optimal policy in order to maximize its own accuracy. This will explain a common frustrating phenomenon in RL, in which the value estimation's loss decreases while the agent's actual scores do not improve, and may even worsen.\\
In many reinforcement-learning algorithms, the brittleness problem is treated as though caused by the training  process arriving at a local-minimum [1]. Accordingly, the solutions include multiple random initializations, parameters tuning and long training with selective termination, in hope the training process will converge to the global minimum. However, in the value-estimation accuracy domain, which is the actual loss function of the value-estimation function, a boring-areas-preferring-policy, or a "boring policy", may be the true global minimum due to low estimation errors of such a policy, while an interesting-areas-preferring-policy, or an "interesting policy", might be a local minimum. This suggests that sometimes, a reported success in a RL task is actually an arrival at a local minimum.
 
\subsection{Exploration importance}
Exploration is a critical part of many current RL algorithms. As its name suggests, exploration is usually perceived as a way to introduce new opportunities to the agent. However, this work can give new insights about what else exploration does and why it is so important. \\
First, in the Boring Areas Trap section above we mentioned that exiting the trap requires a complementary error, but this is sometimes hard to get, since inside the trap we have only examples from the boring area, which have low variance and hence low error rate. Exploration can add examples of interesting areas regardless of the current policy. This can change $Q_i^t$ to a more accurate one, reverse the order between $Q_i^t$ and $Q_b^t$, and thus can help to get out of the Boring Areas Trap.\\
This is especially important in the early stages of the training process, where the random initialization causes approximately half the agents to be in the Boring Areas Trap. This supports starting the training process with high exploration and use a decaying-epsilon-greedy framework to reduce it over time, as customary. \\
Second, in the Manipulative Consultant section we claimed that the consultant influences the examples on which its own accuracy will be tested. Exploration changes the search areas and chooses action regardless of the consultant advice, so the optimal policy becomes the global minimum. \\
***Although its importance, exploration schemes changes the behavior of the agent. There is a need for a complementary method which can overcome the problems while keeping the original purpose of reward maximization, hence produce a reward-maximizing agent. Symmetric rewards noising might be helpful here, as it can change the variance of the boring area, and the loss of the consultant function, without changing the mean reward for a specific state in the long run.

\subsection{The Adaptive Symmetric Reward Noising scheme}
Inspired by the cognitive experiments of Gureckis et al. [7], where low reward noising helped the participants, we propose reward noising for RL algorithms. Following their work, the noise should be a symmetric one so it would not change the mean reward. However, their naive reward noising have two drawbacks: First, the exact amount of noise needs fine parameter tuning, as too much noise can harm the results [7]. Second, since most of the problem is in the low-variance areas, it is better to add noise mainly to them, since adding noise to already noisy areas can reduce performance. Finally, adding uniform noise would not change the basic problem of variance difference. 
As a possible solution we propose the Adaptive Symmetric Reward Noising (ASRN) scheme, which adds a symmetric noise to the rewards, in an adaptive way, so that low variance areas will be added more reward noise than high variance ones, and the noise magnitude will be determined by the highest variance areas. 
First, we need an estimation to the area's variance. In Q-learning, we assume that a good estimation to the area variance is the update $\upsilon_t$ caused by a single reward to the Q-function:
\begin{equation}
	\upsilon_t=\left|Q_{new}(S_t,a_t)-Q_{old}(S_t,a_t)\right|
\end{equation}
	$$=\left|(1-\alpha)Q(S_t,a_t)+\alpha(r_t+\gamma \max_a Q(S_t,a_t)))-Q(S_t,a_t)\right|$$
$$=\left|\alpha((r_t+\gamma \max_a Q(S_t,a_t))-Q(S_t,a_t))\right|$$\\
During an initial training period, we distribute the collected updates $\{\upsilon_t\}$ to N equally-sized bins $b\in\{1,B\}$. In each bin, we collect the appropriate rewards. Once the initial training period has ended, we estimate the standard deviation of the rewards in each bin b:\\
\begin{equation}
	S_b=STD_{j:e_j\in b}(r_j)
\end{equation}
For each bin we calculate the noise $N_b$ 
\begin{equation}
	N_b=\sqrt{S_{max} ^2-S_b ^2} 
	\quad\mathrm{where}\quad 
	S_{max} = \max_b S_b
\end{equation}  
In the rest of the training process, we find for each reward $r_t$ the appropriate bin \textit{b} according to its update $\upsilon_t$. We then sample a noised reward $r_t'$: 
\begin{equation}
r_t' \stackrel{}{\sim} N(r_t,N_b^2)
\end{equation}
We chose this adaptive noising scheme for two reasons:\\
1.	The noise is symmetric, and hence does not influence the behavior of the mean reward.\\
2. After adding the noise, rewards from bins with low update magnitude  $\upsilon_t$ have the same noise as ones with high update magnitude, thus give boring areas variance which is comparable to the variance in interesting areas.

\begin{figure}
	\centering
	\includegraphics[scale=0.42]{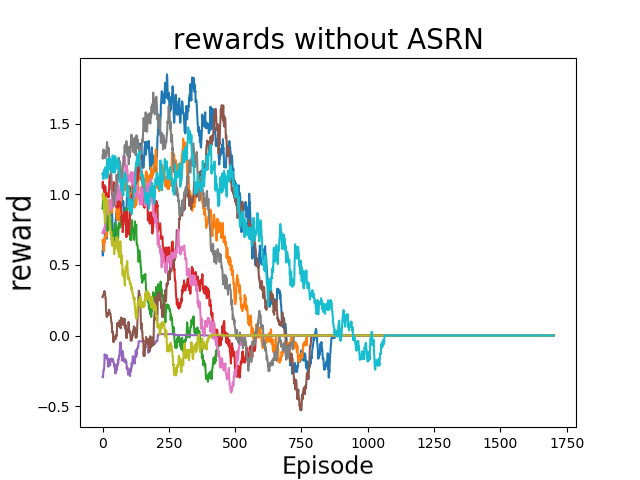}
	\includegraphics[scale=0.42]{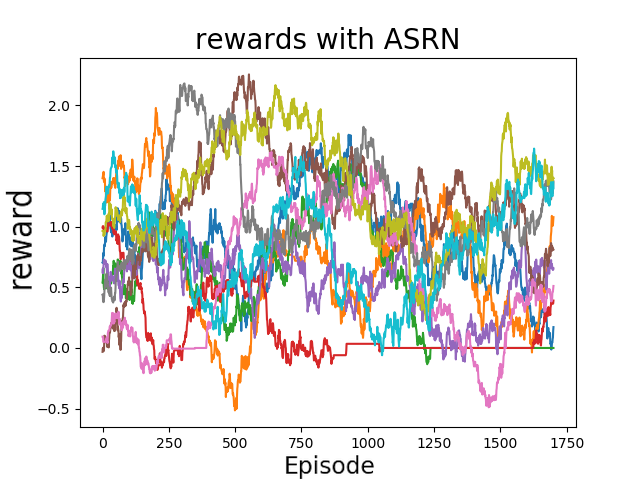}
	\caption{Temporal rewards of ten players training on the Broken-Armed-Bandit game. Rewards were smoothed with a kernel of length 50. All Q-tables were initialized with the optimal value, so all agents start pulling right. \textbf{Left: no ASRN.} All agents move at some point to pull the left arm to gain zero reward, and never come back. \textbf{Right: with ASRN}. The symmetric noise enables the agents to exit the "Boring Areas Trap".}
\end{figure}

\section{Experimental results}
Our first experiment takes place in the well-studied Multi-Armed Bandit environment, with only two arms, with different rewards distributions: the right arm has a positive reward mean and a high reward variance, while the left arm has zero reward mean and low variance. While this problem is easily solved with many basic reinforcement learning algorithms, we found that consultant-based methods like Q-learning show a suboptimal behavior. \\
Since the environment has only one state, the Q-learning update expression is:
\begin{equation}
	Q_{new} (a_t)\leftarrow(1-\alpha)Q(a_t)+\alpha(r_t+\gamma \max_a Q(a))
\end{equation}
The Q-table has only two parameters, $Q_r$,$Q_l$, which are a cumulative-discounted-future-rewards estimation for the right and left arm, respectively. We assume $0<\gamma<1$, and initialize both of them to the optimal values:
\begin{equation}
	Q_r=\sum_{t=0}^\infty\gamma^t\mu_r=\frac{\mu_r}{1-\gamma} 
	\quad\mathrm{and}\quad 
	Q_l=\sum_{t=0}^\infty\gamma^t\mu_l=\frac{\mu_l}{1-\gamma}
\end{equation}
in order to simulate a trained policy and to show the retreat-from-success phenomenon, which we claim causes the brittleness effect. We implemented a decreasing epsilon-greedy exploration framework, with a low decay rate of 0.001, in order to emphasize the limitations of exploration when dealing with the Manipulative Consultant Problem. \\

\subsection{Broken-Armed Bandit demonstrates the Boring Areas Trap}
In this experiment, we ran multiple agents playing for many episodes, with the following parameters: $\mu_l=0$, $\sigma_l=0$, $\mu_r=1.0$, $\sigma_r=2.5$, $\alpha=0.05$, $\gamma=0.95$. We call this extreme example, with zero variance, a "Broken-Armed Bandit", as it simulates a common situation where some action does nothing, or has a deterministic reward behavior.\\  
In Fig. 1 (left) we can see the temporal rewards of ten players during the game. Because all of the agent's Q-function were initialized with the true arms mean reward, they all start with a positive mean return, as they all choose to pull the right arm. However, all of them move at some point to pull the left arm to gain zero reward, and never come back to the right arm. In figure 1 (right) we can see the behavior of the same players after the ASRN scheme added noise to the left arm rewards. All players keep pulling the right arm and get positive rewards, and the brittleness phenomenon disappears. 

\begin{figure}
	\centering
	\begin{minipage}{.48\textwidth}
		\centering
		\includegraphics[width=0.8\linewidth]{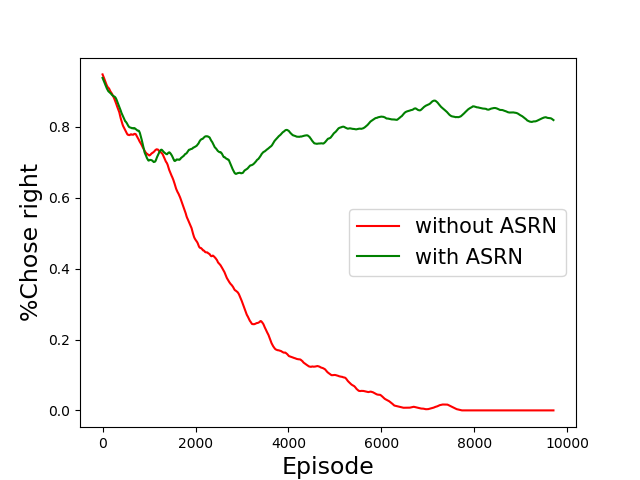}
		\caption{Success probability during a training on the Broken-Armed-Bandit game. \textbf{Without ASRN:} Plain Q-learning algorithm. All agents move eventually to pull the left arm with zero reward, and never recover. \textbf{With ASRN:} Same, with the ASRN scheme. The symmetric noise enables a complementary error, so agents can exit the "Boring Areas Trap".}
		\label{fig:...}
	\end{minipage}%
	\hfill
	\begin{minipage}{.48\textwidth}
		\centering
		\includegraphics[width=0.8\linewidth]{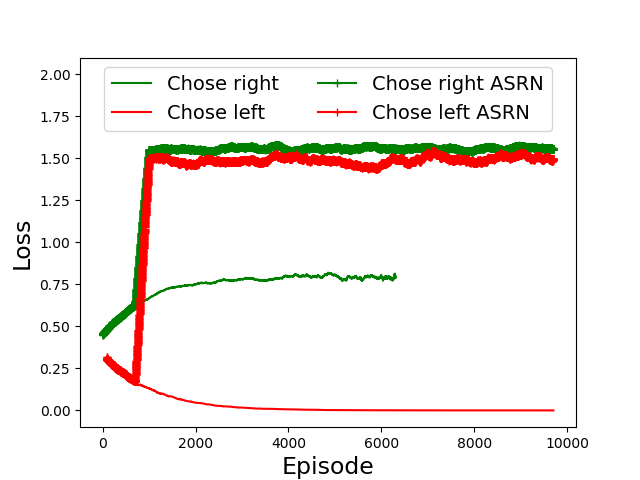}
		\caption{Mean loss during training. \textbf{Chose Right:} currently right-choosing agents. \textbf{Chose left:} currently left-choosing agent. \textbf{Chose right ASRN:} currently right-choosing agent with ASRN. \textbf{Chose left ASRN:} currently right-choosing agent with ASRN. Loss differences are increasing as exploration decays. ASRN equalizes the loss.} 
		\label{fig:...}
	\end{minipage}%
\end{figure}

In Fig. 2 we can see the behavior of 100 players with the same parameters as above, when the graph indicates the number of players currently choosing to pull the right arm. In red we can see the regular players, and in green the players with the ASRN scheme. Eventually, all of the regular players (red) decide to pull the left arm, whereas most of the players with ASRN (green) decide to pull the right arm. This percentage is increasing slowly, meaning there is a way to go back from choosing left arm to choosing right arm and get positive mean rewards. 

%\begin{figure}
%	\centering
%	\includegraphics[scale=0.1]{fig3}
%\end{figure}

\subsection{The Manipulative consultant demonstration}
In Fig. 3 we examine the update term of the Q-learning, $\upsilon_t$, which in DQN methods will be the network loss term, averaging over 100 players with the same parameters as above. When we presented the "Manipulative Consultant" problem we claimed that variance differences causes the loss in boring areas to be lower than the loss in interesting areas, and that this in turn causes the value-estimation function to prefer boring areas. We can see that without the ASRN scheme, the mean loss of policies that chose pulling the right arm (green thin graph)  – which is the interesting area – is indeed higher than of those which chose pulling the left arm (red thin graph) – the boring area. The difference increases as the exploration of the epsilon-greedy method decreases and policies get the reward of their actual decision. This means that a boring policy will show a converging loss over time, only due to the exploration decay, while showing no improvement in rewards terms. In comparison, we can see that when we use the ASRN scheme, after the initial learning period, the loss is similar in both arms, and the "Manipulative Consultant" problem cannot take effect. Note that since we do not optimize a deep-neural-network using this loss, the loss changes does not affect the policy, and this is only a demonstration of their existence.

\subsection{The need for complementary error}
To better understand the Boring Areas Trap phenomenon, we look at one training agent without the ASRN scheme, without exploration and with the following parameters:
$\mu_l=0$, $\sigma_l=0.5$, $\mu_r=1.0$, $\sigma_r=7$, $\alpha=0.1$, $\gamma=0.9$.
In Fig. 4 we examine the parameters $Q_r^t$, $Q_l^t$, which are the only entries of the Q table, along the training process. We can see the policy entering the Boring Areas Trap at episode 40, when it falsely estimates that $Q_r^t<Q_l^t$, and exiting the trap at episode 320, when the complementary error occurs and the policy estimates that  $Q_r^t>Q_l^t$. Estimation of an area does not change when it is not chosen, due to lack of examples. Hence it is clear that exiting the trap has lower probability when the left arm has low variance. 

%\begin{figure}
%	\centering
%	\includegraphics[width=\textwidth]{fig4}
%	\caption{...}
%\end{figure}
 
\begin{figure}
	\centering
	\begin{minipage}{.48\textwidth}
		\centering
		\includegraphics[width=0.8\linewidth]{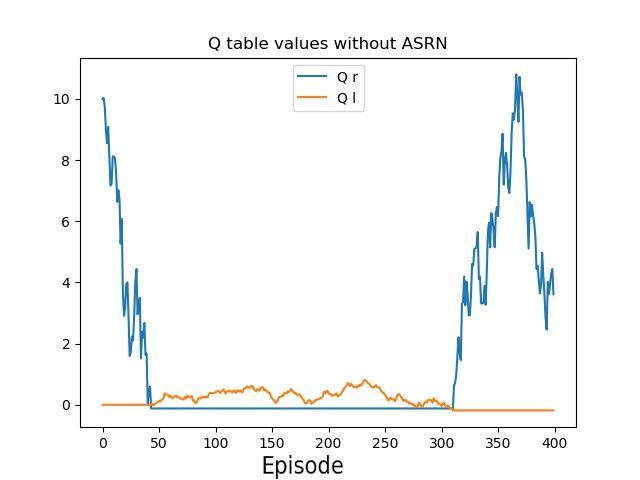}
		\caption{The parameters $Q_r^t$, $Q_l^t$ of a training agent during training. The agent enters the Boring Areas Trap at episode 40, and exits at episode 320, when the complementary error occurs.}
		\label{fig:...}
	\end{minipage}%
	\hfill
	\begin{minipage}{.48\textwidth}
		\centering
		\includegraphics[width=0.8\linewidth]{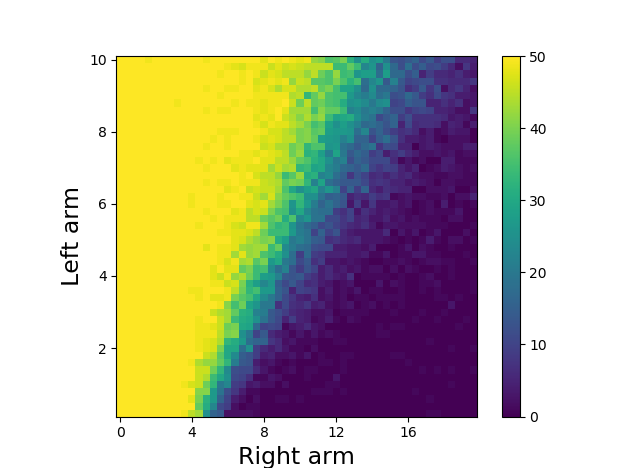}
		\caption{The influence of $\sigma_l$, $\sigma_r$ on the success of 50 agents. \textbf{Right arm:} $\sigma_r$ of right arm. \textbf{Left arm:} $\sigma_l$ of left arm.  }
		\label{fig:...}
	\end{minipage}%
\end{figure} 

\subsection{Different variance values influence}
To see the influence of $\sigma_l$, $\sigma_r$ on the success frequency, we ran a training process with 50 optimally-initialized agents for many combinations of $\sigma_l$ and $\sigma_r$ for 10000 episodes ($\mu_l=0.0$, $\mu_r=1.0$, $\alpha=0.1$, $\gamma=0.9$), and checked how many of them chose the right arm at the last episode. In fig. 5 we can see the results: when the two arms have approximately the same STD, all agents choose the right arm. When $\sigma_l<<\sigma_r$, most agents fail and choose the left arm. In the middle, there is an area in which some of the agents choose right and some choose left. This is similar to the behavior we described in the "Boring Areas Trap" section above. Note that we chose a large learning rate $\alpha=0.1$, and lower $\alpha$ will move the border to the right, so failure will occur only with larger variance differences or smaller mean differences. The code to the first experiment can be found at: \url{www.github.com/ManipulativeConsultant/two-armed-bandit}

\subsection{Autonomous driving using the AirSim simulator}
The second experiment tests the influence of the ASRN scheme in the more complex task of autonomous driving in a driving simulator, using a DQN algorithm. We use the "Neighborhood" environment of the AirSim realistic driving simulator [11], to simulate a car driving in a suburban area, using only one rendered visual image as input per episode, as illustrated in Fig. 6. We follow the end to end deep reinforcement learning algorithm from [12] and the supplied code from [13], which use a deep Q-learning algorithm. We did not use the distributed computing option. We chose to use a transfer learning scheme and to initialize the agent with the provided weights acquired from training the agent on the open-area "Landscape" environment, using supervised learning, in order to accelerate the training process. all experiments were conducted on a HP ZBook 15 G3 computer, with Intel Xeon v5 CPU, 32MB RAM and NVIDIA Quadro GPU. We stopped the training after 2500 games - aproximately 9 hours. We integrated our ASRN scheme implementing the Q-learning version from equations 1-4 above. The integrated code can be found in \url{https://github.com/ManipulativeConsultant/AutonomousDrivingCookbook}\\
In fig. 7 there is a comparison of a training process with and without the ASRN scheme. The graph measures the longest run, in seconds, in the last 20 games, and is smoothed by a low-pass-filter of size 30. We can see that once its training period has ended in game 1000, the ASRN scheme contributes to the car driving duration consistently, while uniform noise reduces them. An example to a drive of the best trained model with ASRN can be found at \url{https://youtu.be/aoft3T_77sQ}

\begin{figure}
	\centering
	\begin{minipage}[t]{0.48\textwidth}
		\includegraphics[width=\textwidth]{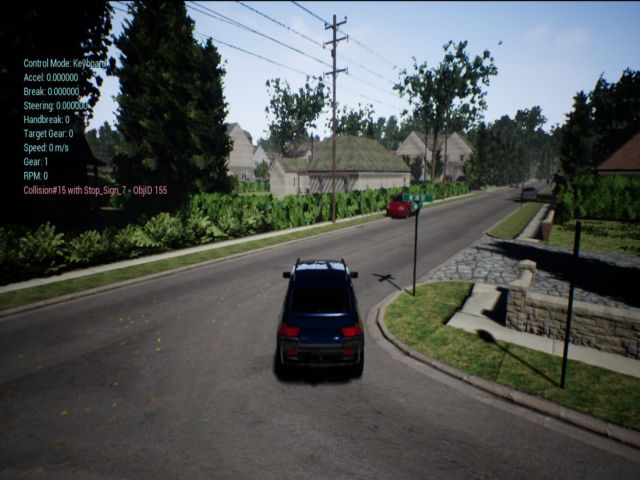}
		\caption{The AirSim realistic driving simulator, with the suburban "Neighborhood" environment. A drive of our best trained model can be found at \url{https://youtu.be/aoft3T_77sQ}}
		\label{fig:...}
	\end{minipage}%
	\hfill
	\begin{minipage}[t]{0.48\textwidth}
		\includegraphics[width=\textwidth]{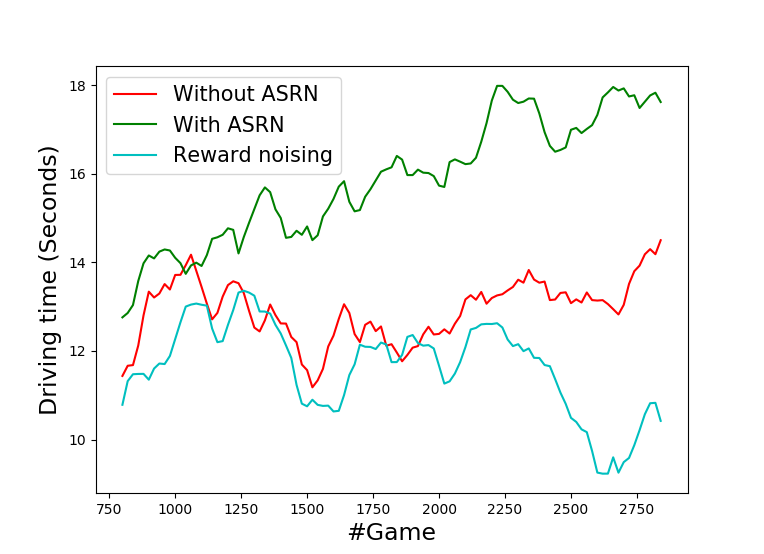}
		\caption{Training DQN agent for self driving with AirSim. Driving time - best from last 20. \textbf{Without ASRN:} DQN algorithm from [12]. \textbf{Reward noising:} with uniform noise of $\sigma=0.1$. \textbf{With ASRN:} using the ASRN scheme.}
		\label{fig:...}
	\end{minipage}%
\end{figure}

\section{Conclusions}
We addressed the brittleness problem in deep reinforcement learning training, and especially the retreat-from-success problem of some algorithms, and the phenomenon where the loss function of some networks continue improving while the performance worsens. We claim that some of the blame should be apportioned to the environment property of variance differences between different areas, and more specifically from the existence of boring areas with low mean reward and low variance. This causes the "Boring Areas Trap" problem, where an agent that decides to do boring actions has low probability of changing its mind because of the low variance. It also causes the "Manipulative Consultant" problem, where the behavior of the value-estimation function converges into suggesting boring policies with low mean reward so as to reduce its own loss function.
We conducted several experiments with a Two-Armed-Bandit environment with variance differences. Although this problem has many excellent RL solutions, we showed that in some cases Q-learning shows the brittleness effect. We also demonstrate the "Manipulative Consultant" problem, where boring policies get better loss values, and the "Boring Areas Trap" where getting out of a boring-policy behavior requires a complimentary error which is rare in low-variance areas. \\
Inspired by the cognitive experiments in [7], suggesting that low reward noising can improve performance, we introduced the Adaptive Symmetric Reward Noising (ASRN) scheme, which adds a normal noise to rewards, adaptively to the amount of this reward estimated variance, so that their long-term mean value stays unchanged while equalizing the noise across the environment. We showed that this can solve the Two-Armed Bandit environment even with severe variance differences. We tested the ASRN contribution to the more complex task of end-to-end autonomous driving, and showed that although we added seemingly misleading noise to the rewards, the performance improved, resulting in longer driving periods. \\
Although we have focused on Q-learning, we believe that recent actor-critic RL algorithms such as A2c and PPO algorithms suffers from the same "Manipulative Consultant" problem, even though the critic has to manipulate the actor network in order to reduce its own loss. Our current noising scheme has several limitations. First, it is designed for Q-learning based algorithms and should be adjusted for actor critic algorithms, maybe by dedicating a separate value-estimation network exclusive for variance estimation. Second, it requires a representative training period, so it must have some exploration to represent many areas. Is assumes a trained agent, trained enough for its estimations to reflect states variance. It also assumes the game does not change after the training period. Finally, the training period is a parameter which should be adjusted to any specific task. In the future we plan to address all of those limitations, conduct a large-scale experiments on rich environments, and check the variance differences influence on other RL algorithms.

\section*{References}

\medskip
\small

[1] Plappert, M., Houthooft, R., Dhariwal, P., Sidor, S., Chen, R.Y., Chen, X., Asfour, T., Abbeel, P. and Andrychowicz, M., (2017) {\it Parameter space noise for exploration.}, arXiv preprint arXiv:  7,
pp.\ 1706.01905 .

[2] Sutton, R. S., Barto, A. G. (1998) {\it  Reinforcement learning: an introduction.} MIT Press. Cambridge, MA.

[3] Pathak, D., Agrawal, P., Efros, A. A., Darrell, T. (2017) {\it Curiosity-driven exploration by self-supervised prediction.} In Proceedings of the IEEE Conference on Computer Vision and Pattern Recognition Workshops (pp. 16-17).

[4] Haarnoja, T., Zhou, A., Abbeel, P., Levine, S. (2018) {\it Soft actor-critic: Off-policy maximum entropy deep reinforcement learning with a stochastic actor}. arXiv preprint arXiv:1801.01290.

[5] https://youtu.be/4gO7uemm6Yo?t=23

[6] Fox, R., Pakman, A., Tishby, N. (2015) {\it  Taming the noise in reinforcement learning via soft updates}. arXiv preprint arXiv:1512.08562.

[7] Gureckis, T.M., Love, B.C. (2009) {\it  Learning in noise: Dynamic decision-making in a variable environment}. Journal of Mathematical Psychology, 53(3), 180-193.

[8] Kaelbling L. P.,  Littman M. L.,  Moore A. W. (1996) {\it Reinforcement learning: A survey}. Journal of artificial intelligence research, 4, 237-285.

[9] Schulman J., Levine S., Abbeel P., Jordan M., Moritz P. (2015) {\it Trust region policy optimization.} In International Conference on Machine Learning (pp. 1889-1897).

[10] Schulman J., Wolski F., Dhariwal P., Radford A., Klimov O. (2017) {\it Proximal policy optimization algorithms.} arXiv preprint arXiv:1707.06347.

[11] Shah S., Dey D., Lovett C., Kapoor A. (2018) {\it Airsim: High-fidelity visual and physical simulation for autonomous vehicles.} In Field and service robotics (pp. 621-635). Springer, Cham.

[12] Spryn, M., Sharma A., Parkar D., Shrimal M. (2018)  {\it Distributed deep reinforcement learning on the cloud for autonomous driving.} In 2018 IEEE/ACM 1st International Workshop on Software Engineering for AI in Autonomous Systems (SEFAIAS) (pp. 16-22). IEEE.

[13] https://github.com/Microsoft/AutonomousDrivingCookbook/tree/master/DistributedRL

\end{document}